\pdfoutput=1

\documentclass[11pt]{article}

\usepackage{acl}

\usepackage{times}
\usepackage{latexsym}

\usepackage[T1]{fontenc}

\usepackage[utf8]{inputenc}
\usepackage{algorithm2e}
\usepackage{booktabs}
\usepackage{microtype}

\usepackage{hyperref}
\usepackage{url}

\usepackage{color}
\usepackage{colortbl}
\usepackage{xcolor}

\usepackage{multirow}
\usepackage{tcolorbox}
\usepackage{graphicx}
\usepackage{subcaption}
\usepackage{amsfonts}
\usepackage{soul}
\usepackage{amsmath}
\usepackage{makecell}
\usepackage{bm}

\newcommand{\ie}{\textit{i.e.}}

\newcommand{\wrt}{\textit{w.r.t}}
\newcommand{\etc}{\textit{etc}}

\makeatletter
\def\@fnsymbol#1{\ensuremath{\ifcase#1\or \dagger \or  \ddagger\or
   \mathsection\or  \text{*}\or \mathparagraph \or  \| \or **\or \dagger\dagger
   \or \ddagger\ddagger \else\@ctrerr\fi}}
\makeatother
 
\renewcommand{\thefootnote}{\fnsymbol{footnote}}

\title{LLMLingua-2: %
Data Distillation for Efficient and Faithful \\
Task-Agnostic Prompt Compression}

\author{
  Zhuoshi Pan$^{1}$\footnotemark[1], Qianhui Wu$^{2}$\footnotemark[2], Huiqiang Jiang$^{2}$, Menglin Xia$^{2}$, Xufang Luo$^{2}$, Jue Zhang$^{2}$,\\
  \textbf{Qingwei Lin$^{2}$, Victor R\"{u}hle$^{2}$, Yuqing Yang$^{2}$, Chin-Yew Lin$^{2}$,} \\
  \textbf{H. Vicky Zhao$^{1}$, Lili Qiu$^{2}$, Dongmei Zhang$^{2}$}\\
  $^{1}$ Tsinghua University, $^{2}$ Microsoft Corporation\\
  \texttt{\{qianhuiwu, hjiang, xufang.luo\}@microsoft.com}\\
}

\begin{document}
\maketitle
\footnotetext[1]{Work during internship at Microsoft.}
\footnotetext[2]{Corresponding author.}

\renewcommand{\thefootnote}{\arabic{footnote}}  %

\begin{abstract}
This paper focuses on task-agnostic prompt compression for better generalizability and efficiency.
Considering the redundancy in natural language, existing approaches compress prompts by removing tokens or lexical units according to their information entropy obtained from a causal language model such as LLaMa-7B.
The challenge is that information entropy may be a suboptimal compression metric: (i) it only leverages unidirectional context and may fail to capture all essential information needed for prompt compression; (ii) it
is not aligned with the prompt compression objective.

To address these issues, we propose a data distillation procedure to derive knowledge from an LLM to compress prompts without losing crucial information, and meantime, introduce an extractive text compression dataset.
We formulate prompt compression as a token classification problem to guarantee the faithfulness of the compressed prompt to the original one,
and use a Transformer encoder as the base architecture to capture all essential information for prompt compression from the full bidirectional context.
Our approach leads to lower latency by explicitly learning the compression objective with smaller models such as XLM-RoBERTa-large and mBERT.

We evaluate our method on both in-domain and out-of-domain datasets, including MeetingBank, LongBench, ZeroScrolls, GSM8K, and BBH. Despite its small size, our model shows significant performance gains over strong baselines and demonstrates robust generalization ability across different LLMs.
Additionally, our model is 3x-6x faster than existing prompt compression methods, while accelerating the end-to-end latency by 1.6x-2.9x with compression ratios of 2x-5x.\footnote{Code: \url{https://aka.ms/LLMLingua-2}}

\end{abstract}

\section{Introduction}
Recent years have witnessed the emergence of various prompting techniques for large language models (LLMs), such as Chain-of-Thought (COT)~\citep{wei2022chain}, In-context Learning (ICL)~\citep{dong2022survey}, and Retrieval Augmented Generation (RAG)~\citep{lewis2020retrieval}.
These techniques empower LLMs to handle complex and varied tasks through rich and informative prompts that may exceed tens of thousands of tokens.
However, the benefits of such lengthy prompts come at a cost of increased computational and financial overhead, as well as the degraded information perception ability of LLMs.
Prompt compression is a straightforward solution to address these issues, which attempts to \textit{shorten the original prompts without losing essential information}.

Several methods have been proposed to compress prompts in a \textit{task-aware} manner
\citep{jiang2023longllmlingua,xu2024recomp,jung2023discrete,huang2023boosting}.
These techniques aim to generate compressed prompts tailored to the specific task or query, typically resulting in enhanced performance on downstream tasks, particularly in question answering.
However, the dependency on task-specific features presents challenges in terms of efficiency and generalizability when deploying these methods.
For example, in RAG-style applications, it may become necessary to compress the same documents multiple times depending on the associated queries with task-aware prompt compression.
More details are discussed in Sec.~\ref{sec:related_work}.

Some works have explored \textit{task-agnostic} prompt compression methods for better generalizability and efficiency \citep{jiang2023llmlingua,li2023compressing}.
The underlying assumption is that \textit{natural language contains redundancy \citep{shannon1951prediction} that may be useful for human understanding but might not be necessary for LLMs.}
Therefore, they propose to compress prompts by removing tokens \citep{jiang2023llmlingua} or lexical units \citep{li2023compressing} according to their information entropy obtained from a causal small language model (SLM), regardless of the downstream task or question information.
However, these task-agnostic methods face two challenges:
(i) Information entropy is an empirical metric for prompt compression. Relying on it for prompt trimming may be suboptimal, as it is not aligned with the prompt compression objective.
(ii) Causal LMs only leverage unidirectional context, which may fail to capture all essential information needed for prompt compression within the context.

The challenges lead to the following research questions:

\paragraph{Q1.} How can we identify or build a suitable dataset to align the SLM towards effective prompt compression?

\paragraph{Q2.} How can we design a compression algorithm that effectively leverages the full bidirectional context for better performance?
\vspace {0.6\baselineskip}

For Q1, most text compression datasets are \textit{abstractive} \citep{toutanova2016dataset,koupaee2018wikihow,kim2019abstractive},
meaning that they treat prompt compression as a generative task where the original prompts are rephrased into condensed ones.
However, this autoregressive generation process is slow and it may produce hallucinated content \citep{zhao2020reducing}.
On the other hand, \textit{extractive} compression datasets such as SentComp \citep{filippova2013overcoming} and DebateSum \citep{roush2020debatesum} are usually created for the summarization task and often lack detailed information.
In the case of prompt compression, this will hurt the performance of LLM inference in downstream applications such as QA (see Appendix~\ref{sec:drawback_dataset} for some examples).
Therefore, it is necessary to construct an extractive text compression dataset that retains essential information.

\paragraph{Contributions.} We present this paper to address the above challenges for task-agnostic prompt compression.
We make the following contributions.

\begin{itemize}
    \item We propose a data distillation procedure to derive knowledge from an LLM (GPT-4) to compress the prompts without losing crucial information.
    We introduce an extractive text compression dataset, containing pairs of original texts from MeetingBank \citep{hu2023meetingbank} and their compressed versions.
    We publicly release the dataset.

    \item We approach prompt compression as a token classification task (\ie, preserve or discard), and take the predicted probability of each token being labeled as \texttt{preserve} as the compression metric.
    The benefits are three folds:
    (1) It can capture all essential information needed for prompt compression from the full bidirectional context by using a Transformer encoder for feature extraction.
    (2) It can lead to lower latency, due to the use of smaller models to explicitly learn the compression objective.
    (3) It guarantees faithfulness of the compressed prompt to the original content.
    
    \item We conduct extensive experiments and analysis on both in-domain (\ie, MeetingBank) and out-of-domain datasets (\ie, LongBench, ZeroScrolls, GSM8K, and Big Bench Hard). Despite small in size, our model shows significant performance gains over strong baselines and demonstrates robust generalization ability from GPT-3.5-Turbo to Mistral-7B. Additionally, our model is 3x-6x faster than existing prompt compression methods, while accelerating the end-to-end latency by 1.6x-2.9x with compression ratios of 2x-5x.

\end{itemize}

\section{Related Works}
\label{sec:related_work}

Depending on whether task information is used for compression, prompt compression methods can be categorized into task-aware and task-agnostic compression approaches.

Task-aware compression compresses the context based on the downstream task or the current query. For example, LongLLMLingua~\citep{jiang2023longllmlingua} applies a question-aware coarse-to-fine compression approach to estimate the information entropy of the tokens and adapts the estimation according to the question. Reinforcement Learning (RL) based methods~\citep{jung2023discrete, huang2023boosting} usually train a model for prompt compression with reward signals from downstream tasks. Soft prompt tuning methods~\citep{wingate-etal-2022-prompt, mu2023learning} typically require fine-tuning for the specific task.
\citet{xu2024recomp} trains a summarization model to compress the context depending on the question. Task-aware compression approaches are usually tailored for specific tasks and compression ratios, which may limit their generalizability in real-world applications.
\begin{figure*}
    \centering    
    \includegraphics[width=2\columnwidth]{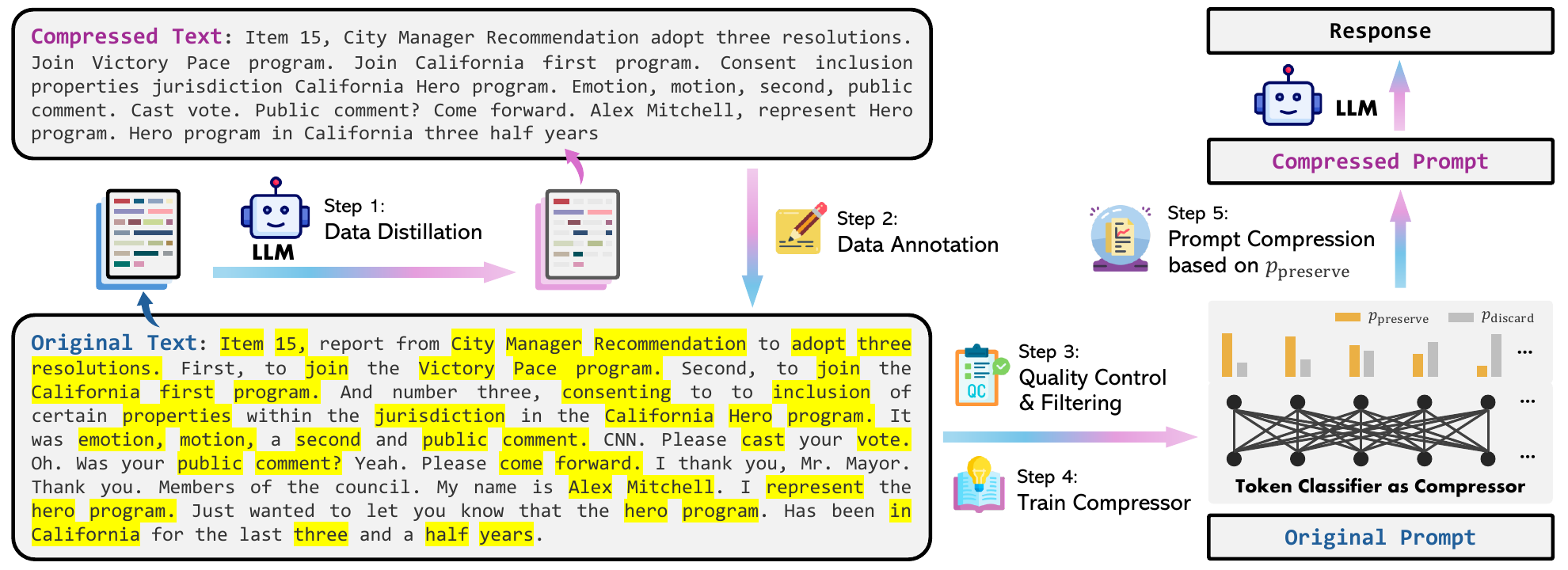}
    \caption{Overview of LLMLingua-2.}
    \label{fig:enter-label}
\end{figure*}

Task-agnostic methods compress the prompt without considering the specific task, making it more adaptable to a range of applications and black-box LLMs. However, producing compressed text that can generalize well to different tasks is not trivial. Typical methods involve using information entropy-based metrics to remove redundant information in the prompt~\citep{li2023compressing, jiang2023llmlingua}. They employ a small language model to estimate token importance from the information metrics. Despite being training-free, these methods may not effectively capture the token importance distribution optimized for specific LLMs and often entail high computation overhead. Summarization-based methods are also leveraged for task-agnostic compression~\citep{chen2023walking, packer2023memgpt}. However, they often omit crucial details and do not generalize well. An alternative approach is to compress or trim the context hidden or KV caches~\citep{chevalier2023adapting, ge2023incontext, zhang2023ho, liu2023scissorhands, xiao2024efficient}. However, this is orthogonal to our work and cannot be easily applied to black-box LLMs.

\section{Dataset Construction}
\label{sec:dataset_construction}

In this section, we outline the process of dataset construction for prompt compression. We first introduce our data distillation procedure, which involves extracting knowledge from an LLM (GPT-4
) to compress texts without losing crucial information or introducing hallucinated content (Sec.~\ref{sec:data_distillation}).
Leveraging the distilled knowledge from the LLM, we explain our data annotation algorithm, which assigns labels to each word in the original text to indicate whether it should be preserved after compression (Sec.~\ref{sec:data_annotation}).
To ensure the dataset's quality, we propose two quality control metrics for filtering low-quality samples (Sec.~\ref{sec:quality_control}).

\subsection{Data Distillation}
\label{sec:data_distillation}

To extract knowledge from the LLM for effective prompt compression, our goal is to prompt GPT-4 to generate compressed texts from original texts that meet the following criteria:
(i) \textit{Token reduction}: Compressed prompts should be short in length to reduce cost and speed up inference.
(ii) \textit{Informativeness}: Essential information should be retained.
(iii) \textit{Faithfulness}: Compressed prompts should remain faithful and avoid introducing hallucinated content to ensure accuracy when prompting LLMs in downstream tasks.

However, distilling such data from GPT-4 is challenging, as it does not consistently follow the instructions.
For instance, \citet{jiang2023llmlingua} experimented with different prompts for compression and found that GPT-4 struggles to retain essential information from original texts. In our preliminary experiments, we have also observed that GPT-4 tends to modify expressions used in the original texts and sometimes generates hallucinated content.
To address this challenge, we propose the following dataset distillation procedure.

\begin{figure}[tb]
    \small
    \begin{tcolorbox}[left=3pt,right=3pt,top=3pt,bottom=3pt]
    \textbf{Our Instruction for Compression:}\\ %
    Compress the given text to short expressions, and such that you (GPT-4) can reconstruct it as close as possible to the original. Unlike the usual text compression, I need you to comply with the 5 conditions below: \\
    1. You can ONLY remove unimportant words. \\
    2. Do not reorder the original words. \\
    3. Do not change the original words. \\
    4. Do not use abbreviations or emojis. \\
    5. Do not add new words or symbols. \\
    Compress the origin aggressively by removing words only. Compress the origin as short as you can, while retaining as much information as possible. If you understand, please compress the following text: 
    \{\textit{text to compress}\}
    The compressed text is: 
    
    \end{tcolorbox}
    \caption{Our instruction used for data distillation.}
    \label{fig:compression_instruction}
\end{figure}

\begin{figure}[tbp]
    \centering
    \includegraphics[width=\linewidth]{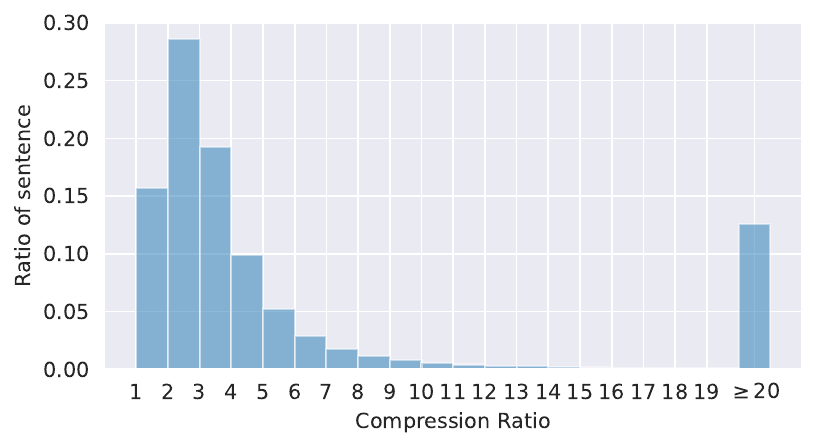}
    \caption{Distribution of compression ratio after chunk-wise compression on MeetingBank.}
    \label{fig: comp_rate_dist_sentence}
\end{figure}

\paragraph{Instruction Design} A well-crafted instruction is the key to unveiling the compression capabilities of GPT-4. 
To ensure that the generated texts stay \textit{faithful} to the original, we explicitly instruct GPT-4 to compress the text by discarding unimportant words in the original texts only and not adding any new words during generation.

To ensure \textit{token reduction} and \textit{informativeness},
previous studies \citep{jiang2023llmlingua, huang2023boosting} have specified either a compression ratio or a target number of compressed tokens in the instructions. However, GPT-4 often fails to adhere to these restrictions.
Additionally, the information density of text can vary significantly depending on its genre, style, etc.
For instance, news articles typically contain denser information compared to meeting transcripts.
Furthermore, even within the domain of meeting transcripts, the information density from different speakers may vary.
These factors suggest that a fixed compression ratio may not be optimal.
Therefore, we remove the compression ratio restriction from our instructions and instead prompt GPT-4 to compress the origin text as short as possible while retaining as much information as possible.
As shown in Fig.~\ref{fig: comp_rate_dist_sentence}, GPT-4 assigns varying compression ratios to different sentences and discards some sentences entirely.
For a comparison between our instruction and those of \citet{jiang2023llmlingua}, please refer to Table~\ref{tab: prompt_analysis}.

\begin{figure}[!t]
\centering
\includegraphics[width=\linewidth]{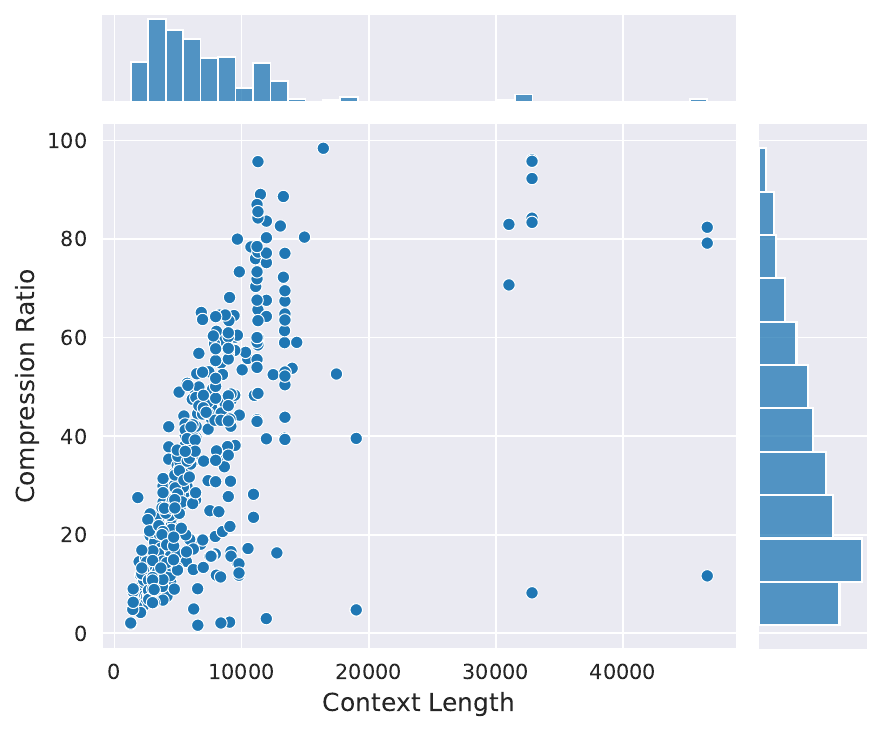}
\caption{
Illustration of compression ratio \wrt. original context length on MeetingBank. We use GPT-4-32k with the output token limit setting to 4096. 
}
\label{fig: comp_ratio_vs_length}
\end{figure}

\paragraph{Chunk-Wise Compression}
Empirically, we have found that the length of the original text has a notable influence on the compression performance.
As shown in Fig.~\ref{fig: comp_ratio_vs_length}, GPT-4 tends to apply a high compression ratio when processing very long context, which might be due to GPT-4's limited ability to handle long context.
This aggressive compression leads to substantial information loss, significantly impacting the performance of downstream tasks.
To mitigate this issue, we first segment each long context into multiple chunks, each containing no more than 512 tokens and ending with a period.
We then instruct GPT-4 to compress each chunk individually.

\subsection{Data Annotation}
\label{sec:data_annotation}
Having obtained pairs of original texts and their compressed versions from data distillation (Sec.~\ref{sec:data_distillation}),
the goal of data annotation is to assign a \textit{binary} label to each token in the original texts to determine if it should be preserved or discarded after compression.
Fig.~\ref{fig: challenges} describes the three primary obstacles encountered here, which arise from GPT-4's inability to precisely comply with the instruction in Fig.~\ref{fig:instruction_us}.
Alg.~\ref{alg:data_annotation} outlines the overall procedure of the proposed annotation algorithm designed to deal with these obstacles.
For more detailed information, please refer to Appendix~\ref{sec:appendix_data_annotataion}.

\begin{figure}[htb]
    \begin{tcolorbox}[left=3pt,right=3pt,top=3pt,bottom=3pt]
    \small
    
    \textbf{Original Texts}\\
    Item 15, report from City Manager Recommendation to adopt three resolutions. First, to join the Victory Pace \textcolor{red}{program}. Second, to join the California first \textcolor{red}{program}. And number three, \hl{consenting} to to \textcolor{blue}{inclusion} of certain \textcolor{blue}{properties} within the jurisdiction in the California Hero \textcolor{red}{program}.
    \vspace{2mm}
    
    \textbf{Compressed Texts}\\
    City Manager Recommendation adopt three resolutions. Join California first \textcolor{red}{program}.
    \hl{Consent} \textcolor{blue}{properties inclusion} jurisdiction California Hero program.
    
    \end{tcolorbox}
    \caption{Challenges in data annotation. \\(i) \textcolor{red}{Ambiguity}: a word in the compressed texts may appear multiple times in the original content. \\(ii) \hl{Variation}: GPT-4 may modify the original words in tense, plural form, \etc. during compression. \\(iii) \textcolor{blue}{Reordering}: The order of words may be changed after compression.}
    \label{fig: challenges}
\end{figure}

\RestyleAlgo{ruled}
\SetKwComment{Comment}{/* }{ */}
\begin{algorithm}[hbt!]
\caption{Data Annotation}
\label{alg:data_annotation}
\SetKwInOut{Input}{Input}
\SetKwInOut{Output}{Output}
\Input{original string $S_{ori}$, compressed string $S_{comp}$, window size $s$.}
Split original string $S_{ori}$ to word list $\mathbb{S}_{ori}$.\\
Split compressed $S_{comp}$ to word list $\mathbb{S}_{comp}$.\\
Initialize labels of original words to \textit{False}. \\
Initialize previous match index $prev$ to $0$.
\For{$w \in \mathbb{S}_{comp}$}{
    \For{$i = 1, 2, ..., \frac{s}{2}$}{
        $right$ = min($|\mathbb{S}_{ori}|, prev + i$) \\
        \If{fuzzy\_match(w, $\mathbb{S}_{ori}[right]$)}{
            $\mathbb{L}$[$right$] = \textit{True}. \\
            $prev$ = $right$.\\
            Break.\\
        }
        $left$ = max($0, prev - i$) \\
        \If{fuzzy\_match(w, $\mathbb{S}_{ori}[left]$)}{
            $\mathbb{L}[left]$ = \textit{True}. \\
            Break.\\
        }
    }
}

\Output{labels of original words $\mathbb{L}(\mathbb{S}_{ori})$.}
\end{algorithm}

\subsection{Quality Control}
\label{sec:quality_control}
We introduce two quality control metrics 
to assess the quality of the compressed texts generated by GPT-4 distillation, as well as the quality of the automatically annotated labels. We then filter the examples by their scores.

\paragraph{Variation Rate}
As GPT-4 may fail to follow the instructions,
we introduce the metric \textit{Variation Rate (VR)} to evaluate the quality of the compressed texts generated from data distillation. VR
measures the proportion of words in the compressed text that are absent in the original text.
Specifically, let $\mathbb{S}_{comp}$ be the set of words in the compressed text and $\mathbb{S}_{ori}$ be that of the original text.
VR is defined as:
\begin{equation}
    \textit{VR} = \frac{1}{|\mathbb{S}_{comp}|} \sum_{w \in \mathbb{S}_{comp}} \mathbb{I} (w \notin \mathbb{S}_{ori}), 
\end{equation}
where $|\cdot|$ is the cardinality of a set.
A higher variation rate implies a higher likelihood of encountering hallucinated content.
Therefore, we exclude the examples with the top 5\% highest variation rates.

\paragraph{Alignment Gap}
We propose \textit{Alignment Gap (AG)} to evaluate the quality of the automatically annotated labels.
Let $l(\cdot)$ represent the annotation function, where $l(w)=\textit{True}$ signifies that word $w \in \mathbb{S}_{ori}$ corresponds to a word in $\mathbb{S}_{comp}$.
We firstly define the matching rate (MR) as:
\begin{equation}
    \textit{MR} = \frac{1}{|\mathbb{S}_{ori}|} \sum_{w \in \mathbb{S}_{ori}} \mathbb{I} (l(w) = \textit{True}).
\end{equation}
Since there exists a many-to-one word mapping from $\mathbb{S}_{ori}$ to $\mathbb{S}_{comp}$ (\ie, the "Ambiguity" challenge presented in Sec.~\ref{sec:data_annotation}), we further present a hitting rate (HR) as a regularization term to measure the proportion of words in $\mathbb{S}_{comp}$ that are found in $\mathbb{S}_{ori}$.
HR is defined as:
\begin{equation}
    \textit{HR} = \frac{1}{|\mathbb{S}_{ori}|} \sum_{w \in \mathbb{S}_{comp}} \mathbb{I} (w \in \mathbb{S}_{ori}).
\end{equation}
Finally, the Alignment Gap (AG) is defined as:
\begin{equation}
    \textit{AG} = \textit{HR} - \textit{MR}.
\end{equation}
The alignment gap of a perfect annotation should be 0.
A large AG indicates a high hitting rate with a poor matching rate, implying low-quality annotation for this example.
Therefore, we discard examples of the highest 10\% alignment gap to ensure quality control of the dataset.

\begin{table*}[tb]
    \centering
    \resizebox{1.9\columnwidth}{!}{
    \begin{tabular}{l| c | c c c c c | c c}
    \toprule
    
    \multirow{2}{*}{\textbf{Methods}}
    & \textbf{QA} & \multicolumn{5}{c|}{\textbf{Summary}}
    & \multicolumn{2}{c}{\textbf{Length}} \\
    \cmidrule (lr){2-2}     \cmidrule (lr){3-7}     \cmidrule (lr){8-9}
    & EM & BLEU & Rouge1 & Rouge2 & RougeL & BERTScore & Tokens & $1/\tau$ \\
    \midrule
    
    Selective-Context & 66.28 & 10.83 & 39.21 & 18.73 & 27.67 & 84.48 & 1,222 & 2.5x  \\
    
    LLMLingua & 67.52 & 8.94 & 37.98 & 14.08 & 26.58 & 86.42 & 1,176 & 2.5x \\
    
    \textbf{LLMLingua-2-small} & \underline{85.82} & \textbf{17.41} & \underline{48.33} & \textbf{23.07} & \textbf{34.36} & \textbf{88.77} & 984 & 3.0x \\

    \textbf{LLMLingua-2} & \textbf{86.92} & \underline{17.37} & 
    \textbf{48.64} & \underline{22.96} & 
    \underline{34.24} & \underline{88.27} & 
    970 & 3.1x \\
    
    \midrule
    Original & 87.75 & 22.34 & 47.28 & 26.66 & 35.15 & 88.96 & 3,003 & 1.0x \\
    
    \bottomrule%
    \end{tabular}
    }
    \caption{In-domain evaluation of different methods on MeetingBank.
    }
    \label{tab:main_meetingbank}
\end{table*}

\section{Compressor}
\label{sec:compression_model}
We formulate prompt compression as a binary token classification problem (\ie, preserve or discard) to guarantee the faithfulness of the compressed prompt to the original content, and meantime ensure the low latency of the compression model itself.
For the token classification model, we employ a Transformer encoder as the feature extractor to leverage information from the bidirectional contexts of each token.
We train the classification model on the dataset constructed in Sec.~\ref{sec:dataset_construction}
from MeetingBank \citep{hu2023meetingbank}.
During inference, we determine whether to preserve or discard each token in the original prompt based on its probability calculated by our classification model.

\subsection{Token Classification Model}
\paragraph{Architecture} We utilize a Transformer encoder \citep{devlin2018bert} as the feature encoder $f_{\theta}$ and add a linear classification layer on top.
Given an original prompt consisting of $N$ words $\bm{x} = \{x_i\}_{i=1}^N$, this can be formulated as:
\begin{align}
    \bm{h} &= f_{\theta}(\bm{x}), \\
    p(x_i, \Theta) &= \text{softmax}(W h_i + b),
\end{align}
where $\bm{h}=\{h_i\}_{i=1}^N$ denotes feature vectors for all words,
$p(x_i, \Theta)\in\mathbb{R}^2$ denotes the probability distribution of labels $\{$\texttt{preserve}, \texttt{discard}$\}$ for the $i$-th word $x_i$,
and $\Theta=\{\theta, W, b\}$ represent all the trainable parameters.

\paragraph{Training} Let $\bm{y}=\{y_i\}_{i=1}^N$ denote the corresponding labels for all words in $\bm{x}$, then we employ cross entropy loss to train the model. The loss function $\mathcal{L}$ \wrt. $\bm{x}$ is:
\begin{equation}
\mathcal{L}(\Theta) = \frac{1}{N}\sum_{i=1}^N \text{CrossEntropy}(y_i, p(x_i, \Theta)).
\end{equation}

\subsection{Compression Strategy}
Our approach to compressing the original prompt $\bm{x}=\{x_i\}_{i=1}^N$ with a target compression ratio 
$1/\tau$ involves a three-step process, where $\tau$ is defined as the quotient of the number of words in the compressed prompt and the number of words in the original prompt $\bm{x}$.
First, we derive the target number of tokens to be preserved in the compressed prompt $\tilde{\bm{x}}$: $\tilde{N} = \tau N$.
Next, we use the token classification model to predict the probability $p_i$ of each word $x_i$ being labeled as \texttt{preserve}\footnote{
To address tokenization-related challenges that arise when applying our approach across various LLMs and SLMs, we preserve the integrity of multi-token words and represent the probability of a word by averaging over the predicted probabilities of all subword tokens.}.
Finally, we retain the top $\tilde{N}$ words in the original prompt $\bm{x}$ with the highest $p_i$ and maintain their original order to form the compressed prompt $\tilde{\bm{x}}$.

It's worth noting that our approach can be readily integrated into the coarse-to-fine framework proposed in LLMLingua \citep{jiang2023llmlingua}, allowing for a higher compression ratio of $\sim$15x for tasks involving multiple demonstrations or documents.
Particularly, we can replace the perplexity-based iterative token compression module in LLMLingua with our token-classification-based compressor, while keeping the budget controller unchanged. Detailed information can be found in Appendix~\ref{sec: integration_with_longllmlingua}.

\begin{table*}[t]
    \centering
    \setlength{\tabcolsep}{1mm}
    \resizebox{2.05\columnwidth}{!}{
    \begin{tabular}{l|ccccccccc|ccc}
    \toprule
        \multirow{2}{*}{\textbf{Methods}} &  \multicolumn{9}{@{}c|}{{\bf LongBench}} &  \multicolumn{3}{@{}c}{{\bf ZeroSCROLLS}} \\
        \cmidrule (lr){2-10} \cmidrule (lr){11-13}

        & SingleDoc & MultiDoc & Summ. & FewShot & Synth. & Code & {\cellcolor[rgb]{0.925,0.957,1}}\textbf{AVG} & Tokens & $1/\tau$ &{\cellcolor[rgb]{0.925,0.957,1}}\textbf{AVG} & Tokens & $1/\tau$ \\

    \midrule
    \midrule
    \multicolumn{6}{@{}r}{{ \textit{2,000-token constraint}}} \\
    \midrule
    \multicolumn{13}{@{}l}{{ \textit{Task(Question)-Aware Compression}}} \\ 
    SBERT$^{\dag}$ & 33.8 & 35.9 & 25.9 & 23.5 & 18.0 & 17.8 & {\cellcolor[rgb]{0.925,0.957,1}}25.8 & 1,947 & 5x & {\cellcolor[rgb]{0.925,0.957,1}}20.5 & 1,773 & 6x \\
    OpenAI$^{\dag}$ & 34.3 & 36.3 & 24.7 & 32.4 & 26.3 & 24.8 & {\cellcolor[rgb]{0.925,0.957,1}}29.8 & 1,991 & 5x & {\cellcolor[rgb]{0.925,0.957,1}}20.6 & 1,784 & 5x \\
    LongLLMLingua$^{\dag}$ & 39.0 & 42.2 & 27.4 & 69.3 & 53.8 & 56.6 & {\cellcolor[rgb]{0.925,0.957,1}}48.0 & 1,809 & 6x & {\cellcolor[rgb]{0.925,0.957,1}}32.5 & 1,753 & 6x \\
    
    \midrule
    \multicolumn{7}{@{}l}{{ \textit{Task(Question)-Agnostic Compression}}} \\
    Selective-Context$^{\dag}$ & 16.2 & \textbf{34.8} & 24.4 & 15.7 & 8.4 & 49.2 & {\cellcolor[rgb]{0.925,0.957,1}}24.8 & 1,925 & 5x & {\cellcolor[rgb]{0.925,0.957,1}}19.4 & 1,865 & 5x \\
    LLMLingua$^{\dag}$ & 22.4 & 32.1 & \underline{24.5} & 61.2 & 10.4 & 56.8 & {\cellcolor[rgb]{0.925,0.957,1}}34.6 & 1,950 & 5x & {\cellcolor[rgb]{0.925,0.957,1}}27.2 & 1,862 & 5x \\
    \textbf{LLMLingua-2-small} & \underline{29.5} & {32.0} & \underline{24.5} & \underline{64.8} & \textbf{22.3} & \underline{56.2} & {\cellcolor[rgb]{0.925,0.957,1}}\underline{38.2} & 1,891 & 5x & {\cellcolor[rgb]{0.925,0.957,1}}\underline{33.3} & 1,862 & 5x \\
    \textbf{LLMLingua-2} & \textbf{29.8} & \underline{33.1} & \textbf{25.3} & \textbf{66.4} & \underline{21.3} & \textbf{58.9} & {\cellcolor[rgb]{0.925,0.957,1}}\textbf{39.1} & 1,954 & 5x & {\cellcolor[rgb]{0.925,0.957,1}}\textbf{33.4} & 1,898 & 5x \\

    \midrule
    \midrule
    \multicolumn{6}{@{}r}{{ \textit{3,000-tokens constraint}}} \\
    \midrule
    \multicolumn{13}{@{}l}{{ \textit{Task(Question)-Aware Compression}}} \\ 
    SBERT$^{\dag}$ & 35.3 & 37.4 & 26.7 & 63.4 & 51.0 & 34.5 & {\cellcolor[rgb]{0.925,0.957,1}}41.4 & 3,399 & 3x & {\cellcolor[rgb]{0.925,0.957,1}}24.0 & 3,340 & 3x \\
    OpenAI$^{\dag}$ & 34.5 & 38.6 & 26.8 & 63.4 & 49.6 & 37.6 & {\cellcolor[rgb]{0.925,0.957,1}}41.7 & 3,421 & 3x & {\cellcolor[rgb]{0.925,0.957,1}}22.4 & 3,362 & 3x \\
    LongLLMLingua$^{\dag}$ & 40.7 & 46.2 & 27.2 & 70.6 & 53.0 & 55.2 & {\cellcolor[rgb]{0.925,0.957,1}}48.8 & 3,283 & 3x & {\cellcolor[rgb]{0.925,0.957,1}}32.8 & 3,412 & 3x \\

    \midrule
    \multicolumn{7}{@{}l}{{ \textit{Task(Question)-Agnostic Compression}}} \\
    Selective-Context$^{\dag}$ & 23.3 & \textbf{39.2} & 25.0 & 23.8 & \textbf{27.5} & 53.1 & {\cellcolor[rgb]{0.925,0.957,1}}32.0 & 3,328 & 3x & {\cellcolor[rgb]{0.925,0.957,1}}20.7 & 3,460 & 3x \\
    LLMLingua$^{\dag}$ & 31.8 & 37.5 & \underline{26.2} & 67.2 & 8.3 & 53.2 & {\cellcolor[rgb]{0.925,0.957,1}}37.4 & 3,421 & 3x & {\cellcolor[rgb]{0.925,0.957,1}}30.7 & 3,366 & 3x \\
    \textbf{LLMLingua-2-small} & \textbf{35.5} & 38.1 & \underline{26.2} & \underline{67.5} & \underline{23.9} & \underline{60.0} & {\cellcolor[rgb]{0.925,0.957,1}}\underline{41.9} & 3,278 & 3x & {\cellcolor[rgb]{0.925,0.957,1}}\underline{33.4} & 3,089 & 3x \\ 
    \textbf{LLMLingua-2} & \textbf{35.5} & \underline{38.7} & \textbf{26.3} & \textbf{69.6} & 21.4 & \textbf{62.8} & {\cellcolor[rgb]{0.925,0.957,1}}\textbf{42.4}  & 3,392 & 3x & {\cellcolor[rgb]{0.925,0.957,1}}\textbf{33.5} & 3,206 & 3x \\
    
    \midrule
    \midrule
    Original Prompt & 39.7 & 38.7 & 26.5 & 67.0 & 37.8 & 54.2 & {\cellcolor[rgb]{0.925,0.957,1}}44.0 & 10,295 & - & {\cellcolor[rgb]{0.925,0.957,1}}34.7 & 9,788 & - \\
    \cmidrule (r){1-1}\cmidrule (lr){2-10} \cmidrule (lr){11-13}
    Zero-Shot & 15.6 & 31.3 & 15.6 & 40.7 & 1.6 & 36.2 & {\cellcolor[rgb]{0.925,0.957,1}}23.5 & 214 & 48x & {\cellcolor[rgb]{0.925,0.957,1}}10.8 & 32 & 306x\\
    \bottomrule
    \end{tabular}
    }
    \caption{Out-of-domain evaluation on general long-context scenarios. $^{\dag}$: numbers reported in \citet{jiang2023longllmlingua}.
    }
    \label{tab:main_longbench}
\end{table*}

\begin{table*}[!ht]
    \centering
    \resizebox{2.05\columnwidth}{!}{
    \begin{tabular}{l|ccc | ccc | ccc | ccc}
    \toprule
        \multirow{3}{*}{\textbf{Methods}} &  \multicolumn{6}{@{}c|}{{\bf GSM8K}} &  \multicolumn{6}{@{}c}{{\bf BBH}} \\
        \cmidrule (lr){2-7}\cmidrule (lr){8-13}
        & \multicolumn{3}{@{}c|}{1-shot constraint} &  \multicolumn{3}{@{}c|}{half-shot constraint} & \multicolumn{3}{@{}c|}{1-shot constraint} &  \multicolumn{3}{@{}c}{half-shot constraint} \\
       \cmidrule (lr){2-4} \cmidrule (lr){5-7}\cmidrule (lr){8-10} \cmidrule (lr){11-13}
        & EM & Tokens & $1/\tau$ & EM & Tokens & $1/\tau$ & EM & Tokens & $1/\tau$ & EM & Tokens & $1/\tau$\\
    \midrule
    Selective-Context$^{\dag}$ & 53.98 & 452 & 5x & 52.99 & 218 & 11x & 54.27 & 276 & 3x & 54.02 & 155 & 5x \\
    LLMLingua$^{\dag}$ & \textbf{79.08} & 446 & 5x & 77.41 & 171 & 14x & \textbf{70.11} & 288 & 3x & \underline{61.60} & 171 & 5x \\
    \textbf{LLMLingua-2-small} & \underline{78.92} & 437 &  5x & \underline{77.48} & 161 & 14x & 69.54 & 263 & 3x & 60.35 & 172 & 5x \\
    \textbf{LLMLingua-2} & \textbf{79.08} & 457 & 5x & \textbf{77.79} & 178 & 14x & \underline{70.02} & 269 & 3x & \textbf{61.94} & 176 & 5x \\
    \midrule
    \midrule
    Full-Shot & 78.85 & 2,366 & - & 78.85 & 2,366 & - & 70.07 & 774 & - & 70.07 & 774 & -  \\
    \midrule
    Zero-Shot & 48.75 & 11 & 215x & 48.75 & 11 & 215x & 32.32 & 16 & 48x & 32.32 & 16 & 48x \\
    \bottomrule
    \end{tabular}
    }
    \caption{Out-of-domain evaluation on reasoning and in-context learning. $^{\dag}$: numbers reported in \citet{jiang2023longllmlingua}.
    }
    \label{tab:main_gsm8k}
\end{table*}

\begin{table*}[tb]
    \centering
    \setlength{\tabcolsep}{1mm}
    \resizebox{2\columnwidth}{!}{
    \begin{tabular}{l|cccc|ccc|ccc}
    \toprule
     \multirow{2}{*}{\textbf{Methods}} &  \multicolumn{4}{@{}c|}{{\bf MeetingBank}} &  \multicolumn{6}{@{}c}{{\bf LongBench-SingleDoc}} \\
    \cmidrule (lr){2-5} \cmidrule (lr){6-11}
    &
    QA & Summ. & Tokens & $1/\tau$ & 
    2,000-token cons. & Tokens & $1/\tau$ &
    3,000-token cons. & Tokens & $1/\tau$  \\
        
    \midrule
    Selective-Context & 58.13 & 26.84 & 1,222 & 2.5x & 22.0 & 2,038 & 7.1x & 26.0 & 3,075 & 4.7x \\
    
    LLMLingua & 50.45 & 23.63 & 1,176 & 2.5x & 19.5 & 2,054 & 7.1x & 20.8 & 3,076 & 4.7x \\
    \textbf{LLMLingua-2-small} &
     \underline{75.97} & 
     \underline{29.93} &                          
    984 & 
    3.0x & 
    
     \underline{25.3} & 
    1,949 & 
    7.4x & 
    
    \textbf{27.9} & 
    2,888 & 
    5.0x \\
    
    \textbf{LLMLingua-2} &
    \textbf{76.22} & 
    \textbf{30.18} & 
    970 & 
    3.0x & 
    
    \textbf{26.8} & 
    1,967 & 
    7.4x & 
    
    \underline{27.3} & 
    2,853 & 
    5.1x \\

    \midrule
        
    Original Prompt & 66.95 & 26.26 & 3,003 & - & 24.5 & 14,511 & - & 24.5 & 14,511 & - \\
    
    \bottomrule
    \end{tabular}
    }
    \caption{Evaluation with Mistral-7B as the Target LLM on MeetingBank and LongBench single doc QA task. We report Rouge1\cite{lin2004rouge} for summary.}
    \label{tab:model_generalization}
\end{table*}

\section{Experiment}
\label{sec:expt}

\paragraph{Implementation Details}
We construct our extractive text compression dataset using training examples from MeetingBank \citep{hu2023meetingbank}
with implementation details in Appendix~\ref{sec:gpt_compression_details}.
Our approach is implemented using Huggingface’s Transformers and PyTorch 2.0.1 with CUDA-11.7.
We use \texttt{xlm-roberta-large} \citep{conneau2019unsupervised} and \texttt{multilingual-BERT} \citep{devlin2018bert} for the feature encoder $f_{\theta}$ in our compressor, which we refer to as \texttt{LLMLingua-2} and \texttt{LLMLingua-2-small}, respectively.
We finetune both models for 10 epochs, using
the Adam optimizer~\citep{kingma2014adam} with a learning rate of 1e-5 and a batch size of 10.
Unless specified otherwise, all reported metrics use GPT-3.5-Turbo-0613\footnote{https://platform.openai.com/} as the target LLM for downstream tasks, with greedy decoding at a temperature of 0 for enhanced stability across experiments.

\paragraph{Datasets \& Evaluation Metrics}
We conduct five groups of experiments to evaluate the compressed prompts on two groups of datasets.

(i) In-Domain: As we train our compressor using the dataset built with training examples from MeetingBank \citep{hu2023meetingbank}, we use the \textbf{MeetingBank} test examples for in-domain evaluation. In addition to the \textit{summarization} task, we further introduce a \textit{QA} task by prompting GPT-4 to generate 3 question-answer pairs for each example distributed across the whole context (see Appendix~\ref{sec: meetingbank_qa} for more details).
For the summarization task, we use the same evaluation metric as in LLMLingua \citep{jiang2023llmlingua}. 
For QA task, we utilize the Exact Match as the evaluation metric. 

(ii) Out-of-Domain: For long-context scenarios, we use \textbf{LongBench} \citep{bai2023longbench} and \textbf{ZeroSCROLLS} \citep{shaham2023zeroscrolls}, and we employ the same evaluation metric as in LongLLMLingua \citep{jiang2023longllmlingua}. For reasoning and in-context learning, we use \textbf{GSM8K} \citep{cobbe2021training} and \textbf{Big Bench Hard (BBH)} \citep{srivastava2023beyond}, with evaluation metrics consistent with LLMLingua \citep{jiang2023llmlingua}.

\paragraph{Baselines} 
We take two state-of-the-art prompt compression methods as primary baselines for comparison: Selective-Context \citep{li2023compressing} and LLMLingua \citep{jiang2023llmlingua}, both are based on \texttt{LLaMA-2-7B}.
Additionally, we compare our approach with some task-aware prompt compression methods, such as retrieval-based methods and LongLLMLingua \citep{jiang2023longllmlingua}.

\paragraph{Results on In-Domain Benchmark}
In Table~\ref{tab:main_meetingbank}, we first present the results of our proposed method compared to the strong baselines on MeetingBank.
Despite the fact that our compressors are much smaller than the LLaMa-2-7B used in the baselines, our approach achieves significantly better performance on both the QA and Summary tasks, and comes close to matching the performance of the original prompt.
This demonstrates the effectiveness of our constructed dataset, and highlights the importance and benefit of optimizing the compression model using prompt compression knowledge.

\paragraph{Results on Out-of-Domain Benchmarks} 
As our model is trained on meeting transcripts data from MeetingBank, here we explore its generalization ability across various benchmarks of long-context scenarios, reasoning, and in-context learning.
Table~\ref{tab:main_longbench} and \ref{tab:main_gsm8k} show the results on LongBench, ZeroSCROLLS, GSM8K, and BBH:
Our model has demonstrated superior performance compared to other task-agnostic baselines.
Even our smaller model, which is of \texttt{BERT-base} size, has been able to achieve comparable, and in some cases, even slightly higher performance than the original prompt.
While our approach has shown promising results, it falls short when compared to other task-aware compression methods like LongLLMlingua \citep{jiang2023llmlingua} on Longbench. We attribute this performance gap to the additional information that they leverage from the question.
However, the task-agnostic characteristics of our model make it an efficient option with good generalizability when deployed across different scenarios.

\paragraph{Mistral-7B as the Target LLM} 
Table~\ref{tab:model_generalization} presents the results of different methods using Mistral-7B-v0.1\footnote{https://mistral.ai/} as the target LLM.
Our method demonstrates significant performance gain over other baselines, showcasing its good generalization ability across target LLMs.
Notably, LLMLingua-2 yields even better performance than the original prompt.
We speculate that Mistral-7B might be less adept at managing long contexts than GPT-3.5-Turbo.
Our method, by offering shorter prompts with higher information density, effectively improves Mistral-7B's final inference performance.

\paragraph{Latency Evaluation}
Table~\ref{tab:latency} shows the latency of different systems on a V100-32G GPU with different compression ratios. It shows that LLMLingua-2 has a much smaller computation overhead than other compression methods, and can achieve an end-to-end speedup ranging from 1.6x to 2.9x.
Additionally, our method can reduce GPU memory costs by 8x, lowering the demand for hardware resources. For details, see the Appendix~\ref{sec:gpu_memory}.

\begin{table}[htb]
    \centering
    \setlength{\tabcolsep}{0.5mm}
    \resizebox{1.03\columnwidth}{!}{
    \begin{tabular}{lcccc}
    \toprule
        $1/\tau$ & 1x & 2x & 3x & 5x \\
         \midrule
        End2End w/o Compression & \multicolumn{4}{c}{14.9} \\
        End2End w/ LLMLingua-2 & - & 9.4 (1.6x) & 7.5 (2.1x) & 5.2 (2.9x) \\
        \midrule
        Selective-Context & - & 15.9 & 15.6 & 15.5 \\
        LLMLingua & - & 2.9 & 2.1 & 1.5 \\
        LLMLingua-2  & - & \textbf{0.5} & \textbf{0.4} & \textbf{0.4} \\
        \bottomrule
    \end{tabular}
    }
    \caption{Latency (s) comparison on MeetingBank. 
    }
    \label{tab:latency}
\end{table}

\paragraph{Observation on Context Awareness} 
We have observed that LLMLingua-2 can effectively maintain the most informative words with respect to the full context as the compression ratio increases.
We owe this to the adoption of the bidirectional context-aware feature extractor, as well as the strategy of explicitly optimizing toward the prompt compression objective.
See Figure \ref{fig: context_aware_compression} for more details.

\begin{table*}[t]
    \centering
    \setlength{\tabcolsep}{1mm}
    \resizebox{2.05\columnwidth}{!}{
    \begin{tabular}{l|ccccccccc|ccc}
    \toprule
    \multirow{2}{*}{\textbf{Methods}} &  \multicolumn{9}{@{}c|}{{\bf LongBench}} &  \multicolumn{3}{@{}c}{{\bf ZeroSCROLLS}} \\
    \cmidrule (lr){2-10} \cmidrule (lr){11-13}

    & SingleDoc & MultiDoc & Summ. & FewShot & Synth. & Code & {\cellcolor[rgb]{0.925,0.957,1}}\textbf{AVG} & Tokens & $1/\tau$ &{\cellcolor[rgb]{0.925,0.957,1}}\textbf{AVG} & Tokens & $1/\tau$ \\
    
    \midrule
    LLMLingua-2-small & 29.5 & {32.0} & 24.5 & 64.8 & 22.3 & 56.2 & {\cellcolor[rgb]{0.925,0.957,1}}38.2 & 1,891 & 5x & {\cellcolor[rgb]{0.925,0.957,1}}33.3 & 1,862 & 5x \\
    LLMLingua-2 & 29.8 & 33.1 & 25.3 & 66.4 & 21.3 & \textbf{58.9} & {\cellcolor[rgb]{0.925,0.957,1}}39.1 & 1,954 & 5x & {\cellcolor[rgb]{0.925,0.957,1}}\textbf{33.4} & 1,898 & 5x \\
    
    LLMLingua-2$^{\ddag}$ & \textbf{30.7} &  \textbf{33.9} & \textbf{25.4} & \textbf{66.6} & \textbf{22.6} & 58.1 & {\cellcolor[rgb]{0.925,0.957,1}}\textbf{39.5} & 1,853 & 5x & {\cellcolor[rgb]{0.925,0.957,1}}\textbf{33.4} & 1,897 & 5x \\

    \midrule
    Original Prompt & 39.7 & 38.7 & 26.5 & 67.0 & 37.8 & 54.2 & {\cellcolor[rgb]{0.925,0.957,1}}44.0 & 10,295 & - & {\cellcolor[rgb]{0.925,0.957,1}}34.7 & 9,788 & - \\
    \cmidrule (r){1-1}\cmidrule (lr){2-10} \cmidrule (lr){11-13}
    Zero-Shot & 15.6 & 31.3 & 15.6 & 40.7 & 1.6 & 36.2 & {\cellcolor[rgb]{0.925,0.957,1}}23.5 & 214 & 48x & {\cellcolor[rgb]{0.925,0.957,1}}10.8 & 32 & 306x\\
    \bottomrule
    \end{tabular}
    }
    \caption{Out-of-domain evaluation on general long-context benchmarks with the 2,000-token constraint. LLMLingua-2$^{\ddag}$: We expand the constructed text compression dataset using 50k examples from TriviaQA-wiki. Then train an LLMLingua-2 compressor with the expanded dataset.
    }
    \label{tab:limitation}
\end{table*}

\begin{table}[!h]
    \centering
    \resizebox{1.03\columnwidth}{!}{
    \begin{tabular}{l|c|c|c}
    \toprule
    \textbf{Instruction} & \makecell{\textbf{$1/\tau$}}& \makecell{\textbf{VR $\downarrow$}} & \makecell{\textbf{QA F1 $\uparrow$}} \\
     \midrule
    Instruction1 & 123x & 13.7 & 19.1 \\
    Instruction2 & 27x & 7.8 & 26.1 \\
    Instruction3 & 78x & 9.6 & 23.7 \\
    Instruction4 & 49x & 9.4 & 24.9 \\
    \midrule
    LLMLingua-2 w/o Chunk & 21x &  \underline{6.0} &  \underline{27.9} \\
    LLMLingua-2 & 2.6x & \textbf{2.2} & \textbf{36.7} \\
    \bottomrule
    \end{tabular}
    }
    \caption{Ablation Study on Chunk-Wise Compression and Instruction Design. We report the compression ratio, variation rate, and QA performance on LongBench Single Document QA. See Fig.\,\ref{fig: other_instructions} in Appendix for more details of Instruction1 - Instruction4 here.
    }
    \label{tab: prompt_analysis}
\end{table}

\paragraph{Prompt Reconstruction}
We have conducted experiments of prompting GPT-4 to reconstruct the original prompt from the LLMLingua-2 compressed prompt.
The results show that GPT-4 can effectively reconstruct the original prompt, suggesting that there is no essential information loss during the compression process of LLMLingua-2.
Figure \ref{fig: prompt_reconstruction} and \ref{fig: prompt_reconstruction2} in Appendix~\ref{sec: prompt_reconstruction} present some examples.

\paragraph{Ablation Study on Chunk-Wise Compression and Instruction Design}
Table \ref{tab: prompt_analysis} shows that both the designed instruction and the chunk-wise compression strategy proposed in this paper significantly contribute to the success of LLMLingua-2.

\section{Conclusion}
\label{sec:conclusion}
This paper targets task-agnostic prompt compression for better generalizability and efficiency.
In this paper, we identify the challenges encountered in existing methods and address them accordingly.
We conduct extensive experiments and analysis on five benchmarks across different tasks and domains.
Our model shows superiority over strong baselines in terms of performance and compression latency.
We publicly release the dataset of text compression with no essential information loss in this paper.

\section*{Limitations}
Our text compression dataset was constructed using only training examples from MeetingBank, a dataset of summarization over meeting transcripts. This raises concerns about the generalization ability of our compressor. Here we discuss this question from two perspectives.

Firstly, we have conducted extensive out-of-domain evaluation on four benchmarks in the paper, including LongBench \citep{bai2023longbench}, ZeroSCROLLS \citep{shaham2023zeroscrolls}, GSM8K \citep{cobbe2021training}, and Big Bench Hard (BBH) \citep{srivastava2023beyond}, which cover multiple tasks from document QA to math problems and in-context learning.
The experimental results show that even our LLMLingua-2-small model that is of \texttt{BERT-base} size achieves superior performance than the two LLaMA-2-7B based baselines Selective-Context \citep{li2023compressing} and LLMLingua \citep{jiang2023llmlingua}.
This demonstrates that our learned prompt compression model has good generalization ability to data from different domains.

Secondly, we expand the constructed text compression dataset using 50k examples from TriviaQA-wiki.
Then train an LLMLingua-2 compressor with the expanded dataset to see whether there would be further performance gain.
Table \ref{tab:limitation} shows the results under the 2,000-token constraint.
We can see that training the compressor with more data does bring further performance gain (LLMLingua-2$^{\ddag}$).
However, the improvement seems not that significant.
We conjecture that this is because although the semantics of texts from different domains may vary a lot,
their redundancy pattern might be similar. Such pattern or knowledge may be learned during in-domain training, and then act as an anchor that can transfer across different domains.
We leave this for future work.

\bibliography{anthology,custom}
\bibliographystyle{acl_natbib}

\appendix

\begin{figure*}[!ht]
    \small
    \begin{tcolorbox}[left=3pt,right=3pt,top=3pt,bottom=3pt]
    \textbf{Prompt Compression Details}:

    \textbf{Example 1}:
    
    \textcolor{gray}{
    \textcolor{red}{Item 15}, report from \textcolor{red}{City Manager} \textcolor{red!33}{Recommendation} to \textcolor{red}{adopt three resolutions.} First, to \textcolor{red}{join} the \textcolor{red}{Victory Pace} \textcolor{red!33}{program.} \textcolor{red!33}{Second}, to \textcolor{red!33}{join} the \textcolor{red}{California} \textcolor{red!66}{first program.} And number \textcolor{red!33}{three}, consenting to to \textcolor{red!66}{inclusion} of certain \textcolor{red}{properties} within the \textcolor{red!33}{jurisdiction} in the \textcolor{red}{California Hero program.} It was \textcolor{red!33}{emotion, motion}, a \textcolor{red!66}{second} and \textcolor{red}{public} \textcolor{red!66}{comment}. \textcolor{red!33}{CNN.} Please \textcolor{red!66}{cast} your \textcolor{red}{vote.} Oh. Was your \textcolor{red!66}{public comment}\textcolor{red}{?} Yeah. Please \textcolor{red!33}{come forward}. I \textcolor{red!33}{thank} you, \textcolor{red!33}{Mr.} \textcolor{red!66}{Mayor.} Thank you. \textcolor{red!66}{Members} of the \textcolor{red!33}{council}. My name is \textcolor{red}{Alex Mitchell}. I \textcolor{red!66}{represent} the \textcolor{red}{hero program}. Just wanted to let you know that the hero program. Has been in \textcolor{red}{California} for the last \textcolor{red}{three} \textcolor{red!33}{and a} \textcolor{red}{half years}. We're in. Over \textcolor{red!66}{20}. We're in \textcolor{red!66}{28} \textcolor{red!33}{counties}, and we've \textcolor{red!66}{completed} over \textcolor{red}{29,000 energy} \textcolor{red!66}{efficient} \textcolor{red}{projects} to make homes. Greener and more energy efficient. And this \textcolor{red!33}{includes} anything. From \textcolor{red!66}{solar} to \textcolor{red!66}{water}. \textcolor{red!33}{Efficiency}. We've done. \textcolor{red!33}{Almost}. \textcolor{red}{\$550 million} in \textcolor{red}{home improvements}. 
    }

    \textbf{Example 2}:
    
    \textcolor{gray}{
    \textcolor{red}{John}\textcolor{red!66}{:} \textcolor{red!33}{So}, um, I've been \textcolor{red!66}{thinking about} the \textcolor{red}{project}, you know, and I \textcolor{red!33}{believe we} \textcolor{red}{need} \textcolor{red!33}{to}, uh, \textcolor{red!66}{make} some \textcolor{red}{changes.} I mean, we \textcolor{red}{want} the \textcolor{red}{project} to \textcolor{red}{succeed}, right\textcolor{red}{?} So, \textcolor{red!33}{like}, I think we \textcolor{red!33}{should} \textcolor{red!66}{consider} maybe \textcolor{red}{revising} the \textcolor{red}{timeline.} \\
    \textcolor{red!66}{Sarah}: I totally \textcolor{red}{agree}, \textcolor{red!66}{John}. I mean, we \textcolor{red!33}{have to be} \textcolor{red}{realistic}, you know. The \textcolor{red}{timeline} \textcolor{red!33}{is}, like, \textcolor{red!66}{too} \textcolor{red}{tight}. You know what I mean\textcolor{red}{?} We \textcolor{red!33}{should} definitely \textcolor{red}{extend} \textcolor{red!33}{it} .
    }
       
    \end{tcolorbox}
    \caption{\textit{LLMLingua-2} performs context awareness compression. The \textcolor{red}{dark red} highlights the words which are preserved at a 5x compression ratio, \textcolor{red!66}{medium red} denotes 3x compression ratio, and \textcolor{red!33}{light red} represents 2x compression ratio. \textcolor{gray}{Gray} indicates discarded words during compression.}
    \label{fig: context_aware_compression}
\end{figure*}

\section{Details of Data Distillation}
\label{sec:gpt_compression_details}
To construct the extractive compression dataset, we use \texttt{GPT-4-32k} to compress the original meeting transcript. Each transcript is divided into chunks first, with each chunk terminating at the end of a complete sentence and not exceeding 512 tokens. We employ the default parameter settings with a temperature of 0.3 and a \texttt{top\_p} of 1.0. The maximum number of generated tokens is set to 4096. Transcripts exceeding 28K tokens are truncated, allowing a 4K token budget for generation. Fig.~\ref{fig:instruction_us} presents the full instruction used in GPT-4 compression. Tab.\,\ref{tab:statistics} shows the statistics of our MeetingBank compression dataset.

\begin{table}[htb]
    \small
    \centering
    \setlength{\tabcolsep}{0.5mm}
    \resizebox{1.0\columnwidth}{!}{
    \begin{tabular}{lcccccc}
    \toprule
    Data Part & Data Size & Chunk  &  Sentence (Avg) & Token (Avg) & $1
    /\tau$ \\
     \midrule
    
    Original & 5,169 & 41,746 & 232 & 3,635 & - \\

    Compressed & 5,169 & 41,746 & 132 & 1,415 & 2.57x \\
    \bottomrule
    
    \end{tabular}
    }
    \caption{
    Statistics of MeetingBank compression dataset.
    }
    \label{tab:statistics}
\end{table}

\section{Details of Data Annotation}
\label{sec:appendix_data_annotataion}
Based on the compressed prompt, we design a word annotation algorithm to automatically assign each word a label indicating whether the word in the original prompt should be retained. Initially, all labels of the original words are set to \textit{False}. Then, for every word in the compressed prompt, we search for its corresponding word in the original prompt, which is then assigned a \textit{True} label.

\paragraph{Sliding Window:} To assign labels to the appropriate words in the original prompt, we utilize a sliding window approach, constraining the search scope within a local window centered on the previously matched word in the original prompt. The search initiates from the last matching position. The \textit{True} label is then assigned to the first matched word in the original prompt. Furthermore, the search is bidirectional to prevent mismatches caused by GPT-4's reordering, as shown in Fig.~\ref{fig: challenges}. Moreover, if GPT-4 introduces new words during compression, the sliding window restricts the search scope, preventing mismatches between the newly added words in the compressed prompt and words in the original prompt.

\paragraph{Fuzzy Matching:} Another challenge arises from the  ``variation" misbehavior of GPT-4, as illustrated in Fig~\ref{fig: challenges}. GPT-4 may alter the original words in tense, voice, and singular/plural forms during compression, even when we request GPT-4 to compress by discarding words only. To address this issue, we first apply lemmatization to reduce words to their base form using Spacy\footnote{https://spacy.io/api/lemmatizer}, and then perform word matching using the sliding window approach.

\section{Context Aware Compression}
\label{sec: context_aware_compression}
Fig.\,\ref{fig: context_aware_compression} presents some compression results of our \textit{LLMLingua-2} under different compression ratios. Our method effectively maintains the most meaningful words as the compression ratio increases.

\section{Comparison with Baselines}
\label{sec: comparison_with_baselines}
In Fig.\,\ref{fig: comparison_with_baseline_bbh} and Fig.\,\ref{fig: comparison_with_baseline_gsm8k}, we qualitatively compare the compressed prompts of our methods with those of baseline method on GSM8K and BBH datasets. Note our \textit{LLMLingua-2} here is only trained on MeetingBank, but also yields more reasonable compressed prompt than baseline methods on the transferred domain data.

\section{Prompt Reconstruction}
\label{sec: prompt_reconstruction}

Fig.\,\ref{fig: prompt_reconstruction} and Fig.\,\ref{fig: prompt_reconstruction2} show two reconstructed prompts from the compressed prompts using GPT-4. Specifically, we prepend a simple reconstruction instruction: "\textit{I have asked you to compress a meeting transcript by dropping word only. Now, reconstruct the original meeting transcript based on the following compressed transcript.}" to the compressed prompt. With the key information preserved in the compressed prompt, the reconstructed prompt closely resembles the original prompt.

\begin{figure*}[htb]
    \small
    \begin{tcolorbox}[left=3pt,right=3pt,top=3pt,bottom=3pt]
    \textbf{Original Prompt (200 Tokens)}: 

    Item 15, report from City Manager Recommendation to adopt three resolutions. First, to join the Victory Pace program. Second, to join the California first program. And number three, consenting to to inclusion of certain properties within the jurisdiction in the California Hero program. It was emotion, motion, a second and public comment. CNN. Please cast your vote. Oh. Was your public comment? Yeah. Please come forward. I thank you, Mr. Mayor. Thank you. Members of the council. My name is Alex Mitchell. I represent the hero program. Just wanted to let you know that the hero program. Has been in California for the last three and a half years. We're in. Over 20. We're in 28 counties, and we've completed over 29,000 energy efficient projects to make homes. Greener and more energy efficient. And this includes anything. From solar to water. Efficiency. We've done. Almost.\$550 million in home improvements. 
    
    \textbf{Compressed Prompt (98 Tokens, 2x Compression)}:
    
    Item 15 report City Manager Recommendation adopt three resolutions. join Victory Pace program. Second join California first program. three inclusion properties jurisdiction California Hero program. emotion motion second public comment. CNN. cast vote. public comment?. come forward. thank you Mr. Mayor. Members council. Alex Mitchell. represent hero program. California three and a half years. 20. 28 counties completed 29,000 energy efficient projects. includes. solar water. Efficiency Almost \$550 million home improvements.
    
    \textbf{Reconstructed prompt (198 Tokens)}:

    Item number 15 on the agenda is the report from the City Manager. The Recommendation is to adopt three resolutions. The first is to join the Victory Pace program. The second resolution is to join the California First program. The third resolution is for the inclusion of properties within our jurisdiction into the California Hero program. There is an emotion for the motion and a second. We will now open the floor for public comment. We have a representative from CNN here. Are there any other individuals who would like to cast a vote or make a public comment? If so, please come forward.

    Thank you, Mr. Mayor and Members of the council. My name is Alex Mitchell and I represent the Hero program in California. I have been with the program for three and a half years. As of now, 20 out of 28 counties have completed 29,000 energy efficient projects. This includes projects related to solar water and energy efficiency. Almost \$550 million has been invested in home improvements.
           
    \end{tcolorbox}
    \caption{Prompt reconstruction results of our \textit{LLMLingua-2} on the MeetingBank example.}
    \label{fig: prompt_reconstruction}
\end{figure*}

\begin{figure*}[!h]
    \small
    \begin{tcolorbox}[left=3pt,right=3pt,top=3pt,bottom=3pt]
    \textbf{Original Prompt (160 Tokens)}:

    Okay. Thank you. Next step is we're going to do item number, is it that was 16. So I could do item 16. We'll try to get through these as expeditiously as possible. And there's going to be a a motion that's ready to go here. So can we the the the item please. Report from city clerk recommendation to receive and file the certification of the petition regarding the regulation of medical marijuana businesses and approve one of the following three alternative actions adopt the initiative ordinance without alteration to submit the initiative ordinance without alteration to the voters to be held on November 8th, 2016 or three. Adopt a report pursuant to California State Elections Code. Thank you. There's a motion and a second device. Marie Lowenthal. Thank you.

    \textbf{Compressed Prompt (83 Tokens, 2x compression)}:

    Okay. Thank you. Next step do item number, 16. item 16. try get through expeditiously. motion ready go here. item. Report from city clerk receive file certification of petition regulation medical marijuana businesses approve three alternative actions adopt initiative ordinance without alteration submit to voters November 8th, 2016 or three. Adopt report California State Elections Code. motion second device. Marie Lowenthal.

    \textbf{Reconstructed Prompt (123 Tokens)}:

    Okay. Thank you. The next step is to do item number 16. For item 16, we will try to get through this expeditiously. The motion is ready to go from here. The item is a report from the city clerk to receive and file the certification of a petition for the regulation of medical marijuana businesses. We can approve three alternative actions: adopt the initiative ordinance without alteration, submit it to the voters on November 8th, 2016, or three, adopt the report as per the California State Elections Code. The motion is seconded by the device. Marie Lowenthal.
    
    \end{tcolorbox}
    \caption{Prompt reconstruction results of our \textit{LLMLingua-2} on the MeetingBank example.}
    \label{fig: prompt_reconstruction2}
\end{figure*}

\begin{figure*}[htb]
    \small
    \begin{tcolorbox}[left=3pt,right=3pt,top=3pt,bottom=3pt]
    \textbf{Our GPT-4 Instruction for Compression:}

    \textbf{System Prompt:}\\
    You are an excellent linguist and very good at compressing passages into short expressions by removing unimportant words, while retaining as much information as possible.
    
    \textbf{User Prompt:}\\
    Compress the given text to short expressions, and such that you (GPT-4) can reconstruct it as close as possible to the original. Unlike the usual text compression, I need you to comply with the 5 conditions below: \\
    1. You can ONLY remove unimportant words. \\
    2. Do not reorder the original words. \\
    3. Do not change the original words. \\
    4. Do not use abbreviations or emojis. \\
    5. Do not add new words or symbols. \\
    Compress the origin aggressively by removing words only. Compress the origin as short as you can, while retaining as much information as possible. If you understand, please compress the following text: 
    \{\textit{text to compress}\}\\
    The compressed text is: 
    
    \end{tcolorbox}
    \caption{The instruction we used in GPT-4 compression.}
    \label{fig:instruction_us}
\end{figure*}

\begin{figure*}[htb]
    \small
    \begin{tcolorbox}[left=3pt,right=3pt,top=3pt,bottom=3pt]
        \textbf{Instruction1:}\\
        Could you please rephrase the paragraph to make it short, and keep 5\% tokens?\\
        \textbf{Instruction2:} \\
        Summarize the provided examples in a few sentences, maintaining all essential reasoning aspects.\\
        \textbf{Instruction3:} \\
        Remove redundancy and express the text concisely in English, ensuring that all key information and reasoning processes are preserved.\\
        \textbf{Instruction4:} \\
        Follow these steps to shorten the given text content: 1. First, calculate the amount of information contained in each sentence, and remove sentences with less information. 2. Next, further condense the text by removing stop words, unnecessary punctuation, and redundant expressions. Refine the content while ensuring that all key information is retained. Let's do it step by step.
    \end{tcolorbox}
    \caption{Other instructions we evaluated, which are proposed in LLMLingua~\citep{jiang2023llmlingua}.}
    \label{fig: other_instructions}
\end{figure*}

\begin{figure*}[htb]
    \small
    \begin{tcolorbox}[left=3pt,right=3pt,top=3pt,bottom=3pt]
    \textbf{Original Prompt (139 tokens)}:

    Q: I have a blackberry, a clarinet, a nectarine, a plum, a strawberry, a banana, a flute, an orange, and a violin. How many fruits do I have? 
    
    A: Let’s think step by step. 
    
    We first identify the fruits on the list and include their quantity in parentheses:
    
    - blackberry (1)
    - nectarine (1)
    - plum (1)
    - strawberry (1)
    - banana (1)
    - orange (1) 
    
    Now, let’s add the numbers in parentheses: 1 + 1 + 1 + 1 + 1 + 1 = 6. So the answer is 6.
    
    \textbf{Compressed prompt (57 tokens) by \textit{LLMLingua}}:

    : a blackberry, a a ne a a a a, many have\\
    :’s think \\
    We first theruits the list and include their in - (– \\
    ’s the numbers in parentheses:1 + 1 = 6. So the answer is 6.
    
    \textbf{Compressed prompt (54 tokens) by \textit{LLMLingua-2}}:

    Q: clarinet, nectarine, strawberry, violin. 
    
    How many fruits 
    
    think step by step.
    
    identify fruits include quantity parentheses: 
    
    blackberry nectarine plum strawberry banana orange add numbers parentheses: 1 + 1 = 6. 
    
    answer is  6.
    
    \end{tcolorbox}
    \caption{Comparison with baseline.  \textit{LLMLingua-2} here is only trained on MeetingBank, but also yields more reasonable compressed prompt than \textit{LLMLingua} on BBH.}
    \label{fig: comparison_with_baseline_bbh}
\end{figure*}

\begin{figure*}[htb]
    \small
    \begin{tcolorbox}[left=3pt,right=3pt,top=3pt,bottom=3pt]
    \textbf{Original Prompt (249 tokens)}:

Question: Sam bought a dozen boxes, each with 30 highlighter pens inside, for \$10 each box. He
rearranged five of these boxes into packages of six highlighters each and sold them for \$3 per
package. He sold the rest of the highlighters separately at the rate of three pens for \$2. How much
profit did he make in total, in dollars? \\
Let’s think step by step \\
Sam bought 12 boxes x \$10 = \$120 worth of highlighters. \\
He bought 12 * 30 = 360 highlighters in total. \\
Sam then took 5 boxes × 6 highlighters/box = 30 highlighters. \\
He sold these boxes for 5 * \$3 = \$15 \\
After selling these 5 boxes there were 360 - 30 = 330 highlighters remaining. \\
These form 330 / 3 = 110 groups of three pens. \\
He sold each of these groups for \$2 each, so made 110 * 2 = \$220 from them. \\
In total, then, he earned \$220 + \$15 = \$235. \\
Since his original cost was \$120, he earned \$235 - \$120 = \$115 in profit. \\
The answer is 115
    
    \textbf{Compressed prompt (144 tokens) by \textit{LLMLingua}}:

: Sam bought a dozen boxes each 30 highl pens inside, \$10 each. He reanged five of boxes into of \\
six each \$3 per. He sold the thelters separately at the of three \$2. much make total, \\
Lets think step \\
bought boxes x0 oflters \\
He 2 3ters in \\
Sam then boxes 6lters/box 0ters \\
He sold these boxes 5 \\
Afterelling these boxes there 36030lters \\
ese00 of three \\
sold groups2 each so made *2 \$20 from \\
In total, he015 \\
Since his he \$ - \$120 = \$115 in profit. \\
The answer is 115
    
    \textbf{Compressed prompt (138 tokens) by \textit{LLMLingua-2}}:

Sam bought dozen 30 highlighter pens \$10 rearranged five boxes into six highlighters sold \$3 per sold rest three pens profit ?\\
 Sam bought 12 boxes x \$10 = \$120 \\
 12 * 30 = 360 highlighters \\
 5 boxes × 6 highlighters/box = 30 \\
 sold 5 * \$3 = \$15 \\
 5 360 - 30 = 330 highlighters \\
 330 / 3 = 110 groups three \\
 sold \$2 110 * 2 = \$220 \\
 earned \$220 + \$15 = \$235.
 original cost earned \$235 - \$120 = \$115 \\
The answer is 115

    \end{tcolorbox}
    \caption{Comparison with baseline. \textit{LLMLingua-2} here is only trained on MeetingBank, but also yields more reasonable compressed prompt than \textit{LLMLingua} on GSM8K.}
    \label{fig: comparison_with_baseline_gsm8k}
\end{figure*}

\section{Details of MeetingBank QA and MeetingBank Summary}
\label{sec: meetingbank_qa}

The MeetingBank QA dataset consists of 862 meeting transcripts from the MeetingBank test set. Initially, we generate 10 question-answer pairs for each meeting transcript using GPT-4-32K.
The instruction used in generating QA pairs is: "\textit{Create 10 questions/answer pairs from the given meeting transcript. The answer should be short and concise. The question should start with Q: and answser should start with A: . The meeting transcript is as follows.}".
To ensure the quality of the generated QA pairs, we discard the question-answer pairs with answer lengths exceeding 50 tokens. Subsequently, we carefully examine the remaining QA pairs to ensure that the answers actually appear in the original transcripts, instead of being products of GPT-4's hallucinations. After the aforementioned filtering process, we retain 3 high-quality question-answer pairs for each meeting transcript. Additionally, we instruct GPT-4-32K to summarize each meeting transcript. The summaries generated by GPT-4 are used as ground truth to evaluate the summary performance.

\section{Drawback of Existing Text Compression Dataset}
\label{sec:drawback_dataset}
Existing extractive compression datasets such as SentComp \citep{filippova2013overcoming} and DebateSum \citep{roush2020debatesum} are mainly created for summarization task. The compressed texts provided in their dataset are usually too concise, only maintaining the main idea of the original text and lacking detailed information. This information loss inevitably hinders the downstream tasks such as document-based QA, as illustrated in Fig.\,\ref{fig: example_of_SentComp} and Fig.\,\ref{fig: example_of_DebateSum} 

\begin{figure*}[htb]
    \small
    \begin{tcolorbox}[left=3pt,right=3pt,top=3pt,bottom=3pt]
    \textbf{Document:}

    \textcolor{blue}{Chinese government is to open more museums, memorial halls} and national patriotism education bases to the public \textcolor{blue}{for free} amid efforts to upgrade cultural services.All national museums and provincial comprehensive museums will stop charging entry fees this year, says a government circular. Museums and memorial halls listed as national patriotism education bases will open for free, adds the circular, jointly issued by the Publicity Department of the Communist Party of China Central Committee, the ministries of finance and culture, and the State Administration of Cultural Heritage on Janyary 23. \textcolor{red}{Free entry is also available to museums above county level in Zhejiang, Fujian, Hubei, Jiangxi, Anhui and Gansu provinces and Xinjiang Uygur Autonomous Region.} Other provinces, autonomous regions and municipalities are encouraged cut or abolish entry fees according to their circumstances, the circular says. All museums, memorial halls and national patriotism education bases will be free to visit by 2009 except cultural relics and historical sites, which will have cheap rates for minors, the elderly, soldiers, the disabled and low-income families, says the circular. For special or guest exhibitions, museums and memorial halls can charge fees, the circular says, and museums are encouraged to have cheap tickets and flexible plans, such as regular free entry, and cheap tickets for groups and families.
    
    \textbf{Question:}
 
    In which provinces will museums above country level be open for free?
    
    \end{tcolorbox}
    \caption{An example from the \textit{SentComp} dataset \citep{filippova2013overcoming}. The compressed text is highlighted in \textcolor{blue}{blue}. The provided compressed text fails to cover the question references which are highlighted in  \textcolor{red}{red}.}
    \label{fig: example_of_SentComp}
\end{figure*}

\begin{figure*}[htb]
    \small
    \begin{tcolorbox}[left=3pt,right=3pt,top=3pt,bottom=3pt]
    \textbf{Document:}

    The overall results regarding the long-term effects of exchange rate volatility are highly informative in relation to the exports and imports of an LDC. \textcolor{blue}{Mexico's exports of agricultural goods are clearly depressed by uncertainty}: Table 3 shows that \textcolor{blue}{no unprocessed agricultural good responds positively, while various animal, vegetable, and wood products make up 6 of the 21 industries with negative effects. Imports are also affected}. While the category of Oil-seeds, oil nuts, and oil kernels does seem to increase because of uncertainty, \textcolor{blue}{6 of the 21 industries in which volatility reduces import flows are agricultural in nature. Mexican textile exports also show clear negative effects due to uncertainty}, \textcolor{red}{not only for the category of Clothing except fur clothing, but also for the inputs of Textile and leather machinery and Textile yarn and thread} (in Table 4).

    \textbf{Question:}
 
    Which industries of textile suffer from negative effects due to the exchange rate uncertainty?
    
    \end{tcolorbox}
    \caption{An example from the \textit{DebateSum} dataset \citep{roush2020debatesum}. The compressed text is highlighted in \textcolor{blue}{blue}. The provided compressed text fails to cover the question references which are highlighted in  \textcolor{red}{red}.}
    \label{fig: example_of_DebateSum}
\end{figure*}

\section{Model Size and Training Details}

We use \texttt{xlm-roberta-large} which has 355M parameters as the feature encoder $f_{\theta}$ in \texttt{LLMLingua-2}. The training
process takes approximately 23 hours on our MeetingBank compression dataset.  For \texttt{LLMLingua-2-small}, the feature encoder is the \texttt{multilingual-BERT} which has 110M parameters. It takes roughly 16 hours to train the \texttt{multilingual-BERT} model.  

\section{GPU Memory Usage}
\label{sec:gpu_memory}

LLMLingua-2 enjoys a smaller GPU memory overhead because of its lightweight. 
The peak GPU memory usage of LLMLingua-2 on MeetingBank is only 2.1GB, while LLMLingua and Selective-Context, which utilize LLAMA-2-7B as the SLM, consume 16.6GB and 26.5GB of GPU memory, respectively.

\section{Multilingual Generalization Ability}
In Table~\ref{tab:main_longbench_zh}, we assess the performance of \textit{LLMLingua-2} on the Chinese benchmarks of LongBench, comprising 5 tasks with a total of 1000 samples. Despite being trained solely on the MeetingBank data, which consists of English corpus only, \textit{LLMLingua-2} also outperforms \textit{LLMLingua} on Chinese benchmarks. 
We attribute this performance gain to the multilingual capabilities of the \texttt{xlm-roberta-large} or \texttt{multilingual-BERT} compressor acquired from the pre-training phase.

\section{Integration with LongLLMLingua}
\label{sec: integration_with_longllmlingua}
In retrieval-augmented generation (RAG) and Multi-Documents Question-Answer (MDQA) scenarios, the primary challenge is to identify the document that contains the key information relevant to the question. In these scenarios, \textit{LongLLMLingua} improves the key information preservation by utilizing the information provided in the question. 

While \textit{LLMLingua-2} is designed for question-agnostic compression, it can also be integrated with \textit{LongLLMLingua} to preserve more key information relevant to the question in these scenarios. 
Specifically, we utilize \textit{LongLLMLingua's} coarse-grained compression to assign varying compression ratios to different documents based on the question's perplexity conditioned on each document. Consequently, it allocates more token budgets to the documents which are more relevant to the question.

As illustrated in Table~\ref{tab:nq_results}, \textit{LLMLingua-2} with \textit{LongLLMLingua} coarse-grained compression achieves an average performance gain of 25.3\% on NaturalQuestions \citep{liu2023lost} compared to \textit{LLMLingua-2}.

\section{Sample-Wise Dynamic Compression Ratio}

By default, \textit{LLMLingua-2} applies fixed compression rate to all samples in the benchmark. However, this approach may not be optimal due to variations in the density of key information across different samples. To address this problem, we allow \textit{LLMLingua-2} to dynamically adjust the compression rate for each sample under the overall compression rate constraint.
Specifically, we employ the compressor to predict each token's preservation probability of all samples. We then set a probability threshold to achieve the overall compression rate constraint. For all samples, tokens with preservation probabilities higher than this threshold are retained.

Table~\ref{tab:sample_wise_dynamic_on_longbench} presents the performance of \textit{LLMLingua-2} using the sample-wise dynamic compression ratio, showcasing a 4.4\% and 4.5\% performance improvement under 7x and 5x compression ratios, respectively, compared to \textit{LLMLingua-2} with a fixed compression ratio.

\section{Performance \textit{w.r.t} Compression Ratio}

\begin{figure*}[htbp]
\centering
\begin{subfigure}{0.45\linewidth}
        \includegraphics[width=\linewidth]{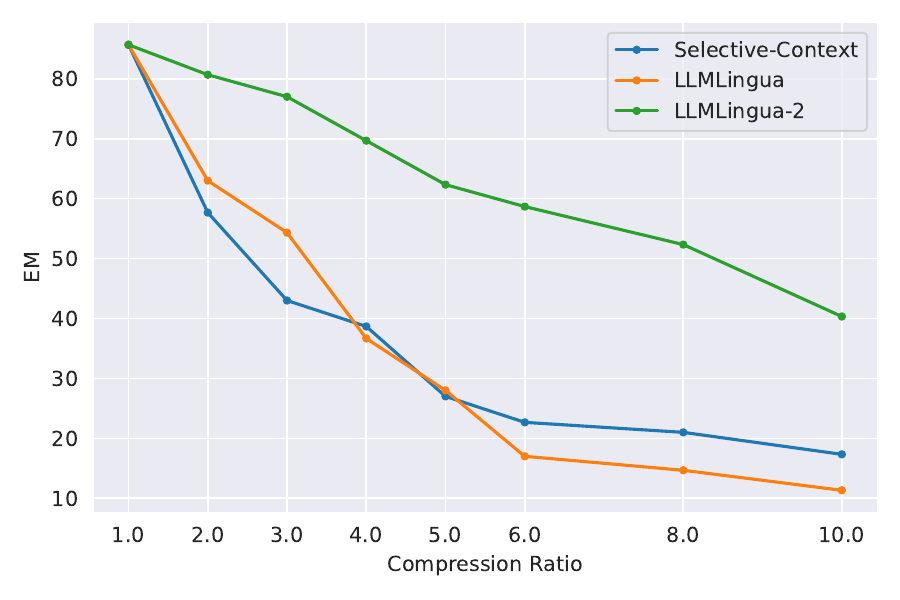}
        \caption*{(a) QA performance \textit{w.r.t} compression ratio on a 100 samples subset of MeetingBank.}
    \end{subfigure}
\hfill
    \begin{subfigure}{0.45\linewidth}
        \includegraphics[width=\linewidth]{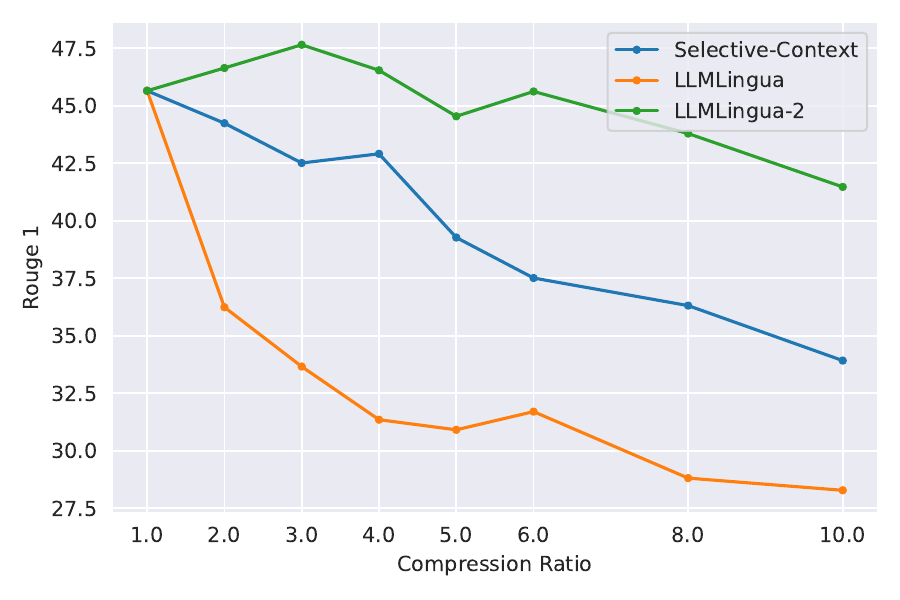}
        \caption*{(b) Summary performance \textit{w.r.t} compression ratio on a 100 samples subset of MeetingBank.}
    \end{subfigure}
\caption{
A plot of performance \textit{w.r.t} compression ratio on a 100 samples subset of MeetingBank.
}
\label{fig: performance_vs_compression_ratio}
\end{figure*}

Fig~\ref{fig: performance_vs_compression_ratio} presents the performance \textit{w.r.t} compression ratio on a subset of 100 samples from Meetingbank. As depicted, LLMLingua-2 exhibits superior robustness compared to other baselines as the compression ratio increases.

\section{Preservation Priority in GPT-4 Compression}

\begin{figure*}[htbp]
\centering
\includegraphics[width=\linewidth]{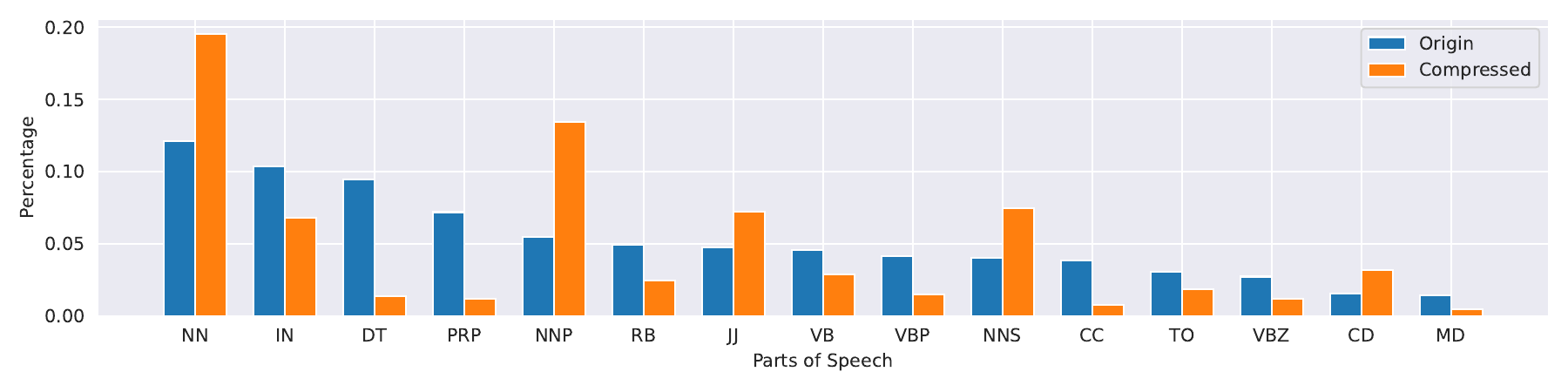}
\caption{
Part of speech distribution of the original prompts and GPT-4 compressed prompts.
}
\label{fig: pos_ratios}
\end{figure*}

To gain insight into GPT-4's compression patterns, we analyze the distribution of different POS categories. Fig~\ref{fig: pos_ratios} suggests that GPT-4 prioritizes the preservation of nouns, adjectives, and numerals, which typically play a more important role in the comprehension of the overall context.

\section{Comparison With GPT-4 Compression}

Table~\ref{tab: ablation_meetingbank} shows the comparison between LLMLingua-2 compressed prompts and GPT-4 compressed prompts. For GPT-4 compression, We use the same compression instruction as the one used in training data collection. The same chunking technique is also adopted with the chunk size setting to 512. It is shown that LLMLingua-2 achieves higher performance than GPT-4 compression on MeetingBank QA.
We conjecture that LLMLingua-2's ability to learn compression knowledge from the entire dataset helps mitigate the influence of noise and information loss present in each GPT-4 compressed example, leading to superior performance.

\begin{table}[tb]
    \centering
    \resizebox{1\columnwidth}{!}{
    \begin{tabular}{l| c | c c}
    \toprule
    
    \multirow{2}{*}{\textbf{Methods}}
    & \textbf{QA} 
    & \multicolumn{2}{c}{\textbf{Length}} \\
    \cmidrule (lr){2-2} \cmidrule (lr){3-4}
    & EM & Tokens & $1/\tau$ \\
    \midrule

    \textbf{GPT-4 Compression} & 84.86 & 
    1,221 & 2.5x \\
    
    \textbf{LLMLingua-2-small} & 85.82 & 984 & 3.0x \\

    \textbf{LLMLingua-2} &  \textbf{86.92} &     970 & 3.1x \\
    
    \midrule
    Original & 87.75 & 3,003 & 1.0x \\
    
    \bottomrule%
    \end{tabular}
    }
    \caption{Comparison with GPT-4 compressed prompt on MeetingBank.}
    \label{tab: ablation_meetingbank}
\end{table}

\section{Performance of Mistral-7B on 8K Token Subset}

As the Mistral 7B model is trained with an 8k context length \footnote{https://huggingface.co/docs/transformers/main/en/model\_ doc/mistral} , its performance may drop if the input context is too long. Therefore, we conduct additional experiments on subsets containing only examples with original prompts shorter than 8k tokens. The results, shown in Table~\ref{tab: longbench_single_doc_qa_8k}, demonstrate that LLMLingua-2 continues to outperform strong baselines and even the original prompts in this subset.

\begin{table*}[t]
    \centering
    \begin{tabular}{l|cccccccc}
    \toprule
        \multirow{2}{*}{\textbf{Methods}} &  \multicolumn{8}{@{}c}{{\bf LongBench-Zh}} \\
        \cmidrule (lr){2-9}

        & SingleDoc & MultiDoc & Summ. & FewShot & Synth. & {\cellcolor[rgb]{0.925,0.957,1}}\textbf{AVG} & Tokens & $1/\tau$ \\

    \midrule
    \multicolumn{7}{@{}l}{{ \textit{Task(Question)-Agnostic Compression}}} \\
    \midrule

    \textbf{LLMLingua} & 35.2 & 20.4 & 11.8 & 24.3 & 51.4 & {\cellcolor[rgb]{0.925,0.957,1}}28.6 & 3,060 & 5x \\
    
    \textbf{LLMLingua-2} & \textbf{46.7} & \textbf{23.0} & \textbf{15.3} & \textbf{32.8} & \textbf{72.6} & {\cellcolor[rgb]{0.925,0.957,1}}\textbf{38.1}  & 3,023 & 5x \\
    
    \midrule
    \midrule
    Original Prompt & 61.2 & 28.7 & 16.0 & 29.2 & 77.5 & {\cellcolor[rgb]{0.925,0.957,1}}42.5 & 14,940 & - \\
    \bottomrule
    \end{tabular}
    \caption{Out-of-domain evaluation on LongBench \textbf{Chinese} benchmarks.}
    \label{tab:main_longbench_zh}
\end{table*}

\begin{table*}[tb]
    \centering
    \begin{tabular}{l|cccccc|cc}
    \toprule
        Methods & 1st & 5th & 10th & 15th & 20th & Reorder & Tokens & $1/\tau$ \\

    \midrule
    \midrule
    \multicolumn{9}{@{}c}{{ \textit{4x constraint}}} \\
    \midrule
    \multicolumn{7}{@{}l}{{ \textit{Question-Aware Compression}}} \\
    
    BM25$^{\dag}$ & 40.6 & 38.6 & 38.2 & 37.4 & 36.6 & 36.3 & 798 & 3.7x \\
    
    Gzip$^{\dag}$ & 63.1 & 61.0 & 59.8 & 61.1 & 60.1 & 62.3 & 824 & 3.6x \\
    SBERT$^{\dag}$ & 66.9 & 61.1 & 59.0 & 61.2 & 60.3 & 64.4 & 808 & 3.6x \\
    OpenAI$^{\dag}$ & 63.8 & 64.6 & 65.4 & 64.1 & 63.7 & 63.7 & 804 & 3.7x \\
    {\cellcolor[rgb]{0.925,0.957,1}}\textbf{LLMLingua-2$^+$} & 
    {\cellcolor[rgb]{0.925,0.957,1}}74.0 & 
    {\cellcolor[rgb]{0.925,0.957,1}}70.4 & 
    {\cellcolor[rgb]{0.925,0.957,1}}67.0 & 
    {\cellcolor[rgb]{0.925,0.957,1}}66.9 & 
    {\cellcolor[rgb]{0.925,0.957,1}}65.3 & 
    {\cellcolor[rgb]{0.925,0.957,1}}71.9 & 
    {\cellcolor[rgb]{0.925,0.957,1}}739 & 
    {\cellcolor[rgb]{0.925,0.957,1}}3.9x \\
    
    {\cellcolor[rgb]{0.925,0.957,1}}\textbf{LongLLMLingua}$^{\dag}$ & 
    {\cellcolor[rgb]{0.925,0.957,1}}\textbf{75.0} & 
    {\cellcolor[rgb]{0.925,0.957,1}}\textbf{71.8} &
    {\cellcolor[rgb]{0.925,0.957,1}}\textbf{71.2} & 
    {\cellcolor[rgb]{0.925,0.957,1}}\textbf{71.2} & 
    {\cellcolor[rgb]{0.925,0.957,1}}\textbf{74.7} & 
    {\cellcolor[rgb]{0.925,0.957,1}}\textbf{75.5} & 
    {\cellcolor[rgb]{0.925,0.957,1}}748 & 
    {\cellcolor[rgb]{0.925,0.957,1}}3.9x \\
    
    \cmidrule (r){1-1}\cmidrule (lr){2-7} \cmidrule (lr){8-9}
    \multicolumn{7}{@{}l}{{ \textit{Question-Agnostic Compression}}} \\
    Selective-Context$^{\dag}$ & 31.4 & 19.5 & 24.7 & 24.1 & 43.8 & - & 791 & 3.7x \\
    LLMLingua$^{\dag}$ & 25.5 & 27.5 & 23.5 & 26.5 & 30.0 & 27.0 & 775 & 3.8x \\

    \textbf{LLMLingua-2} & 
    48.6 & 
    44.5 & 
    43.6 &
    40.9 & 
    39.9 & 
    46.2 & 
    748 & 
    3.9x \\

    \midrule
    \midrule
    Original Prompt & 75.7 & 57.3 & 54.1 & 55.4 & 63.1 & - & 2,946 & - \\
    \cmidrule (r){1-1}\cmidrule (lr){2-7} \cmidrule (lr){8-9}
    Zero-shot & & & \multicolumn{2}{@{}c}{{ 56.1}} & & & 15 & 196x \\
    \bottomrule
    \end{tabular}
    \caption{
    Performance comparison on NaturalQuestions (20 documents)~\citep{liu2023lost}. \textit{LLMLingua-2$^+$} denotes \textit{LLMLingua-2} with \textit{LongLLMLingua} \cite{jiang2023longllmlingua} coarse level compression.
    $^{\dag}$: numbers reported in \citet{jiang2023longllmlingua}.
    }
    \label{tab:nq_results}
\end{table*}

\begin{table*}[tb]
    \centering
        \begin{tabular}{l|ccc|ccc}
        \toprule
        \multirow{2}{*}{\textbf{Methods}} &  \multicolumn{6}{@{}c}{{\bf LongBench-SingleDoc}} \\
        \cmidrule (lr){2-4} \cmidrule (lr){5-7}
        & QA Score & Tokens & $1/\tau$ &
        QA Score & Tokens & $1/\tau$  \\
        \midrule
        \midrule
        \textit{Target Token Constraint} & \multicolumn{3}{@{}c|}{\textit{2,000 Tokens}} & \multicolumn{3}{@{}c}{\textit{3,000 Tokens}} \\
        \midrule
        LLMLingua-2 & 29.8 & 1,954 & 7.4x & 35.5 & 3,392 & 4.3x \\

        \midrule
        \midrule
        \textit{Compression Ratio Constraint} & \multicolumn{3}{@{}c|}{\textit{7x}} & \multicolumn{3}{@{}c}{\textit{5x}} \\
        \midrule
        
        LLMLingua-2 FR$^{\dag}$ & 25.1 & 2,131 & 6.8x & 27.4 & 3,185 & 4.5x \\

        LLMLingua-2 DCR$^{\ddag}$ & \textbf{29.5} & 2,125 & 6.8x & \textbf{32.2} & 3,164 & 4.5x \\
        
        \midrule
        \midrule
        Original Prompt & 39.7 & 14,511 & 1x & 39.7 & 14,511 & 1x \\
        
        \bottomrule
        \end{tabular}
    \caption{Evaluation of LLMLingua-2 sample wise dynamic compression on LongBench single doc QA task. FR$^{\dag}$ assigns each example with the same fixed compression rate. DCR$^{\ddag}$ assigns dynamic compression rate to different examples within the corpus level constraint. }
    \label{tab:sample_wise_dynamic_on_longbench}
\end{table*}

\begin{table*}[tb]
    \centering
    \setlength{\tabcolsep}{1mm}
    \resizebox{2\columnwidth}{!}{
    \begin{tabular}{l|cccc|ccc|ccc}
    \toprule
     \multirow{2}{*}{\textbf{Methods}} &  \multicolumn{4}{@{}c|}{{\bf MeetingBank}} &  \multicolumn{6}{@{}c}{{\bf LongBench-SingleDoc}} \\
    \cmidrule (lr){2-5} \cmidrule (lr){6-11}
    &
    QA & Summ. & Tokens & $1/\tau$ & 
    2,000-token cons. & Tokens & $1/\tau$ &
    3,000-token cons. & Tokens & $1/\tau$  \\
        
    \midrule
    Selective-Context & 62.43 & 19.25 & 703 & 2.4x & 29.3 & 1,829 & 2.5x & 34.6 & 2,855 & 1.6x \\
    
    LLMLingua & 51.78 & 24.57 & 714 & 2.4x & 29.9 & 1,862 & 2.5x & 30.7 & 3,016 & 1.5x \\
        
    \textbf{LLMLingua-2} &
    \textbf{81.75} & 
    \textbf{30.83} & 
    651 & 
    2.6x & 
    
    \textbf{35.0} & 
    1,889 & 
    2.4x & 
    
    \textbf{36.3} & 
    2,841 & 
    1.6x \\

    \midrule
        
    Original Prompt & 71.27 & 27.56 & 1,700 & - & 31.4 & 4,595 & - & 31.4 & 4,595 & - \\
    
    \bottomrule
    \end{tabular}
    }
    \caption{Evaluation with Mistral-7B as the Target LLM on MeetingBank and LongBench single doc QA task. We discarded samples where the input text has more than 8K tokens. We report Rouge1\cite{lin2004rouge} for summary.}
    \label{tab: longbench_single_doc_qa_8k}
\end{table*}

\end{document}